\pdfoutput=1

\documentclass[11pt]{article}

\usepackage[]{EMNLP2023}

\usepackage{times}
\usepackage{latexsym}

\usepackage{adjustbox}
\usepackage{multirow}
\usepackage{amsmath}  

\usepackage{booktabs} 

\usepackage[T1]{fontenc}

\usepackage[utf8]{inputenc}

\usepackage{microtype}

\usepackage{inconsolata}

\title{Towards Multiple References Era - Addressing Data Leakage and Limited Reference Diversity in NLG Evaluation}

\author{
  Xianfeng Zeng,~
  Yijin Liu,~ 
  Fandong Meng\thanks{~ Corresponding author.} ~
  and Jie Zhou \\
  Pattern Recognition Center, WeChat AI, Tencent Inc, China \\
  \texttt{\{xianfzeng, yijinliu, fandongmeng, withtomzhou\}@tencent.com} \\
}

\begin{document}
\maketitle
\begin{abstract}
N-gram matching-based evaluation metrics, such as BLEU and chrF, are widely utilized across a range of natural language generation (NLG) tasks. 
However, recent studies have revealed a weak correlation between these matching-based metrics and human evaluations, especially when compared with neural-based metrics like BLEURT. 
In this paper, we conjecture that the performance bottleneck in matching-based metrics may be caused by the limited diversity of references.
To address this issue, we propose to utilize \textit{multiple references} to enhance the consistency between these metrics and human evaluations.
Within the WMT Metrics benchmarks, we observe that the multi-references F200spBLEU surpasses the conventional single-reference one by an accuracy improvement of 7.2\%. 
Remarkably, it also exceeds the neural-based BERTscore by an accuracy enhancement of 3.9\%.
Moreover, we observe that the data leakage issue in large language models (LLMs) can be mitigated to a large extent by our multi-reference metric.
We release the code and data at \url{https://github.com/SefaZeng/LLM-Ref}
\end{abstract}

\section{Introduction}
Due to the inherent diversity and complexity of natural language, human evaluation serves as the gold standard for assessing the quality of natural language generation (NLG) tasks. 
However, conducting human evaluation is time-consuming and prohibitively expensive in real-world scenarios. Consequently, the development of reliable automatic evaluation metrics is crucial for advancing NLG research and optimizing NLG systems~\citep{celikyilmaz2020evaluation}. 
The ideal automatic evaluation metrics need to assess the accuracy, fluency, fidelity, and diversity (e.g., conversion) of the generated candidates by the model.
It also requires a high degree of consistency with human assessment to demonstrate its reliability.

\begin{figure}[t!]
\begin{center}
     \scalebox{0.9}{
      \includegraphics[width=0.5\textwidth]{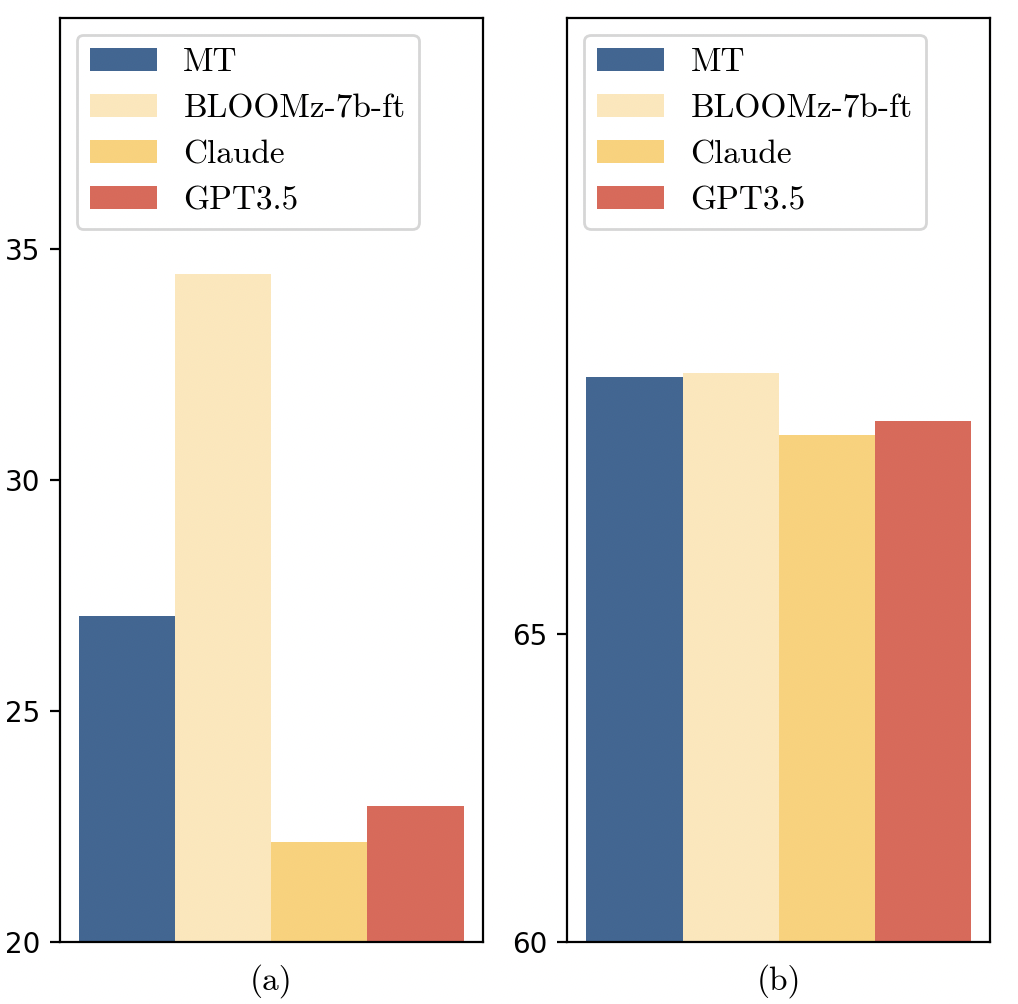}
      } 
      \caption{
      F200spBLEU and BLEURT scores of Japanese to Chinese translation task on Flores200 test set.
      BLOOMz-7b-ft perversely outperforms the strong MT baseline and close source LLMs in F200spBLEU.
      } 
      \label{fig:flores}  
 \end{center} 
\end{figure}

Automatic evaluation metrics for NLG can be generally classified into two categories: N-gram matching based metrics and neural-based metrics. N-gram matching based metrics, {\em e.g.,} BLEU~\cite{papineni2002bleu} primarily calculate the lexical overlap between model's outputs and the ground truth. On the other hand, neural-based metrics {\em e.g.,} BLEURT~\cite{sellam2020bleurt} are trained on a huge amount of text data and score the candidates with a black-box model. 
Due to the huge amount of training data and parameters, neural-based metrics generally achieve better robustness and generalization.
Recent studies~\citep{freitag-etal-2022-results, kocmi-etal-2022-findings} have demonstrated that neural-based metrics exhibit better agreement with human evaluation when compared with N-gram matching based metrics.

The emergence of the Large Language Models (LLMs) has further deepened these concerns~\citep{zhao2023survey} due to the diversity of output results from LLMs. 
LLMs have demonstrated impressive performance across various NLG tasks, leading to a growing trend of fine-tuning LLMs for specific language generation tasks.
Recent studies~\citep{freitag-etal-2022-results} have highlighted the challenges in using N-gram matching based metrics, such as BLEU~\citep{papineni2002bleu}, to evaluate LLM-generated hypotheses. 
Some latest reports~\citep{anil2023palm} exclusively rely on neural-based metrics.

In this paper, we try to investigate why n-gram-based metrics are ineffective to evaluate the quality of LLM-generated candidates and give quantitative results from the perspective of diversity on token distribution.
Then we propose LLM-Ref, a framework that uses LLMs to generate multiple references and select them with high diversity, to improve the accuracy of automatic evaluation metrics.
Experimental results show that our framework improves the consistency between human evaluation and all kinds of automatic evaluation metrics.
Further analysis reveals that multiple references with n-gram-based metrics can effectively mitigate the potential data leakage risk of LLMs which neural-based metrics have difficulty overcoming.

The contributions of this paper can be summarized as follows:

\begin{itemize}
    \item We investigate why n-gram-based metrics, {\em e.g.,} BLEU is ineffective to measure the quality of LLM generation well and provide quantitative results about diversity of references.
    \item We propose LLM-Ref, a framework to generate multiple synthetic references for NLG tasks and conduct diversity-aware filtering.
    \item Our framework improves the consistency between automatic evaluation metrics and human evaluation by a large margin, and achieves state-of-the-art results for non-LLM metrics on WMT22 Metrics Task.    
    \item We further emphasize the necessity of N-gram matching based metrics as they substantially alleviate the miss-evaluation problem due to data leakage risk when multiple references are used while neural-based metrics struggle to overcome.
\end{itemize}

\begin{table}[t]
\begin{center}
\scalebox{0.72}{
\begin{tabular}{ll|c|c}
\toprule
\textbf{ID} & \textbf{Model} & \textbf{DistinctN (n=6) $\uparrow$} & \textbf{Unique Token $\uparrow$} \\ 
\midrule
0 & AISP-SJTU & 0.7399 & 7418 \\
1 & HuaweiTSC & 0.7423 & 7210 \\
2 & M2M100\_1.2B-B4 & 0.7137 & 7023 \\
3 & Online-G & 0.7532 & 7467 \\
4 & Ref-A (Ground Truth) & 0.7520 & 7244 \\
5 & GPT3.5 & \textbf{0.7857} & \textbf{8222} \\
\bottomrule
\end{tabular}
}
\caption{The DistinctN and Unique Token Number for each model's outputs. The token distribution of GPT3.5 is much more diverse than MT systems.}
\label{tab:distinctn}
\end{center}
\end{table}

\section{Preliminary Experiments}
In our preliminary experiments, we conduct supervised fine-tuning (SFT) on the Bloomz~\citep{scao2022bloom} with bilingual corpus in Chinese and Japanese mainly from the CCMatrix dataset~\citep{schwenk-etal-2021-ccmatrix}, and then evaluate on the Flores test set~\citep{costa2022no}. 
To provide a comprehensive comparison, we also evaluate the performance of closed-source large models such as GPT-3.5 and Claude.
The baseline translation model (MT) is an in-house machine translation model with the conventional encoder-decoder architecture.
The evaluation metrics utilized in our experiments included the representative ngram-based BLEU metric~\citep{papineni2002bleu,costa2022no} and the neural-based BLEURT metric~\citep{sellam2020bleurt}. 
The summarized results are presented in Figure \ref{fig:flores}. 
We observe two conclusions from the above experiments:

\begin{itemize}
\item The BLEU scores varies significantly across models, while the BLEURT scores varies relatively flat across models.
\item Bloomz suffers from test data leakage issue, and thus it perversely outperforms the strong MT model.
\end{itemize}

\begin{figure*}[t!]
\begin{center}
     \scalebox{0.9}{
      \includegraphics[width=1\textwidth]{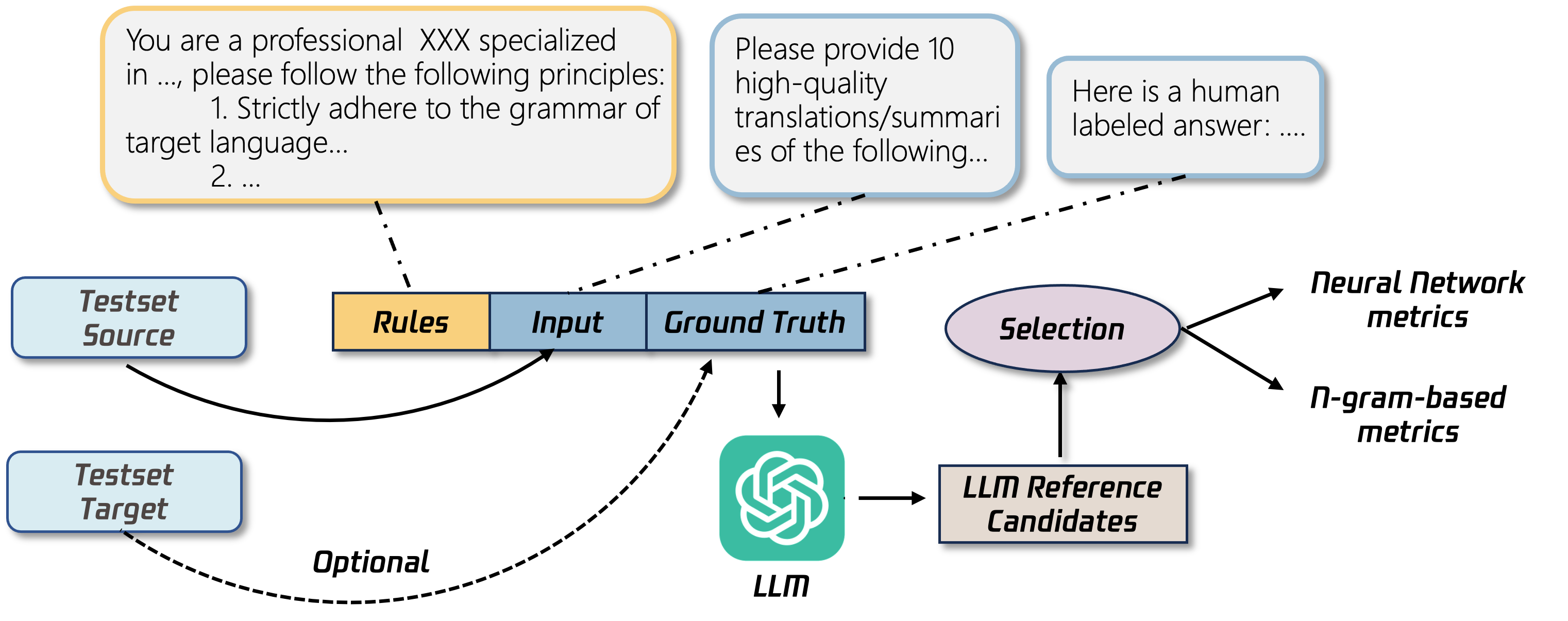}
      } 
      \caption{
      The overall pipeline of LLM-Ref. The prompt consists of Rules, Input, and Ground Truth(optional). Rules contain characterization settings and rules for specific tasks. The input contains the task description and source from the test set. Ground Truth is optional. We generate multiple reference candidates using LLMs and then select the candidates with high diversity.  
      } 
      \label{fig:generation}  
 \end{center} 
\end{figure*}

\paragraph{The Inadequacy of N-Gram-Based Metrics for Evaluating LLM.}
For many years, BLEU~\citep{papineni2002bleu} has been the predominant evaluation metric for machine translation. However, recent studies have revealed that BLEU exhibits weaker agreement with manual evaluation compared to other n-gram-based metrics. Meanwhile, the agreement between neural-based metrics and human evaluation is substantially higher than that of n-gram-based metrics. The advent of LLM has further exacerbated this discrepancy. 
To gain further insights into why BLEU demonstrated such a pronounced bias in evaluating LLM results, we investigated the output distribution of the LLM. We postulate that the output distribution of the LM model differs significantly from that of the MT model, and the n-gram-based nature of BLEU is constrained by the provided reference, resulting in degraded performance when there is a substantial divergence between the output and the reference.

In our experiments, we employed the Chinese to English dataset from the WMT22 Metrics shared task~\cite{freitag-etal-2022-results}. We primarily employed DistinctN~\citep{li2015diversity} and the count of unique tokens in the output to analyze the distribution of the LM model. We selected system outputs and ground truth translations from the Metrics shared task as samples for our analysis. The results are presented in Table \ref{tab:distinctn}.

We discovered that GPT3.5 exhibited higher DistinctN scores than all MT models, including commercial systems, and even outperformed human-translated reference translations. Additionally, the count of unique tokens generated by GPT3.5 was significantly larger than that of other systems. These findings indicate that the output of the LM model exhibits a high degree of diversity. It is the diversity that renders it challenging for n-gram-based metrics, e.g., BLEU, to effectively estimate the quality of generated results.

\paragraph{LLM's Data Leakage Risk.}
The open-source LLMs have gradually become common foundational models for the research community. However, due to the intricate and time-consuming data processing involved in the pre-training phase, the occurrence of data leakage during this phase is common. Recent studies~\citep{zhu2023multilingual} have identified that the BLOOMz model suffers from data leakage when evaluated on the Flores test set, which is a widely adopted large-scale multilingual machine translation test set, thereby hindering its use as an evaluation benchmark for BLOOMz even other LLMs.

Since data leakage can lead to the model generating outputs that closely resemble the reference results, we hypothesize that employing multiple references can alleviate this problem to some extent. Overfitting can occur due to the model's focus on specific references, resulting in poor generalization and suboptimal performance on other references.

\section{Methodology}
To obtain high-quality reference candidates for a more accurate evaluation of the model's generation performance, we employ a series of steps, as outlined below:

\subsection{LLM Reference Candidates Generation}
LLMs demonstrate strong capabilities in a variety of natural language processing tasks. We plan to leverage LLMs' powerful generative capabilities to generate diverse reference candidates which further enhance the NLG evaluations. The prompt for invoking LLMs consists of three components.
\paragraph{Rules.} 
The rules mainly contain some characterization (e.g., You are a professional translator...), and rules to follow. These rules aim to facilitate LLMs generating higher quality and more human-preferred reference candidates.
\paragraph{Input.}
The input consists mainly of a task-specific description (e.g., Please provide 10 high-quality translations/summaries of the following...) and the source of the test set.
\paragraph{Ground Truth.}
Providing manually annotated Ground Truth enables LLMs to generate better and more diverse reference candidates with an awareness of human-preferred answers. The provision of Ground Truth is optional as sometimes we don't always hold high-quality manual references.

The nuances within the prompt can influence the quality of generated candidates, and there are many factors that affect the prompt. We will delve into a detailed analysis of the impact of prompt variants in Section \ref{sec:prompt}.

\subsection{Diversity-Aware Selection}
While the LLMs are capable of generating multiple references, there is a limitation on the number of results that can be formulated from the same meaning. Consequently, as the number of generated reference candidates increases, the language model tends to produce outputs with limited diversity (few word substitutions). Therefore, it is crucial to employ suitable strategies for selection from the generated reference candidates.

We utilize Self-BLEU~\citep{zeng-etal-2021-wechat,meng-etal-2020-wechat} as a metric of diversity for the generated reference candidates. Self-BLEU evaluates the diversity of each reference candidate by computing its BLEU score against the other reference candidates, which is calculated as:

\begin{small}
\begin{equation}
\text{Self-BLEU}_{i} = BLEU(y_i, [y_0, ..., y_{i-1}, y_{i+1}, ...,y_n])
\label{equ:deepnorm_beta}
\end{equation}
\end{small}
Where $y_i$ is the \textit{i}-th reference candidate, and each reference will get a score $\text{Self-BLEU}_{i}$. 
We select the generated reference candidates with Self-BLEU less than 35 which is an empirical number.

\subsection{Multi-Reference for Neural-Based Metrics}
Neural-based metrics primarily assess the model's performance by scoring individual model output in the test set with references and subsequently averaging these scores through the whole test set to get the system-level score for the model. As most neural-based metrics only calculate model output with one reference, we have explored several simple methods(average, top-k, etc..)to combine the scores with multiple references. Empirical results indicate that selecting the maximum value has a positive effect while others have negative impacts. Consequently, all subsequent experiments with neural-based metrics are based on the $max$ method, as specified by the following calculation:

\begin{equation}
\begin{aligned}
\text{scores}&= \{\text{Metric}(\hat{y}, y_0), \text{Metric}(\hat{y}, y_i), ...\} \\
\text{score}&= max(\text{scores}) 
\end{aligned}
\label{equ:nn_metric}
\end{equation}
Here $\hat{y}$ is the output from the model and $y_i$ represents the \textit{i}-th reference candidate generated by large language models and every single sentence in the test set gets a score.

\section{Experiments}
In this section, we describe the benchmarks and the evaluation we used for our experiments. We choose two NLG tasks about machine translation and summarization.

\subsection{Datasets and Evaluation}
In order to evaluate the performance of our proposed framework, we conducted experiments on both WMT22 Metrics shared task \citep{freitag-etal-2022-results} and SummEval benchmark~\citep{fabbri2021summeval}.
\paragraph{WMT22 Metrics Shared Task.} This task includes human judgments for three translation directions: English to German (EN-DE), English to Russian (EN-RU), and Chinese to English (ZH-EN). The three directions consist of 54 machine translation system outputs or human translations, encompassing a total of 106,000 sentences. 

The test set for each direction contains around 2000 sentences covering several text domains. The evaluation criteria are based on MQM\footnote{MQM datasets are composed of multi-dimensional quality scoring by expert translators. Crowdsourced DA datasets are direct assessment scores from crowdsourcing staff.} datasets annotated by domain experts, and related studies have shown that this approach is more accurate than crowdsourced DA datasets~\citep{freitag2021experts}.

To evaluate the correlation between automatic metrics and human evaluation, we measured system-level pairwise accuracy, Pearson correlation (\textbf{$\rho$}), and Kendall-Tau correlation (\textbf{$\tau$}), following the methodology established by the WMT22 Metrics shared task \citep{kocmi-etal-2021-ship}. Pairwise accuracy measures the number of system pairs ranked correctly by the metric compared to the human ranking, divided by the total number of system pair comparisons. It is calculated as follows~\citep{kocmi2023large}:

\begin{small}
\begin{equation}
    \text{Accuracy} = \frac{|sign(metric\Delta) = sign(human\Delta)|}{|\textit{all system pairs}|}
\end{equation}
\end{small}
The Pearson correlation evaluates the linear relationship between automatic metric scores and MQM scores, while the Kendall-Tau correlation is based on pairwise score comparisons and reflects ranking consistency.

We reproduced reported scores in the WMT22 Metrics shared task findings~\citep{freitag-etal-2022-results} using the official WMT22 script.\footnote{https://github.com/google-research/mt-metrics-eval}

\paragraph{SummEval Benchmark.} SummEval comprises 100 summaries generated by each of the 16 models on the CNN/Daily Mail dataset~\citep{see-etal-2017-get}. Human judgments are collected from both experts and crowd-sourced. They assess these summaries in terms of coherence, consistency, fluency, and relevance. 

Following their experiments, we report the sample-level Spearman score~\citep{zar2005spearman} to measure the correlation.

\subsection{Baseline Metrics}
As baseline metrics, we mainly focus on two types of metrics including n-gram-based metrics and neural-based metrics:
\paragraph{N-gram-based metrics.}
N-gram-based metrics include the following:
\begin{itemize}
    \item \textbf{BLEU~\citep{papineni2002bleu}} is based on the precision of n-grams between the MT output and its reference weighted by a brevity penalty.
    \item \textbf{F200spBLEU~\citep{costa2022no}} are BLEU scores computed with subword tokenization done by standardized Sentencepiece Models.
    \item \textbf{CHRF~\citep{popovic-2015-chrf}} uses character n-grams instead of word n-grams to compare the MT output with the reference.
    \item \textbf{ROUGE~\citep{lin-2004-rouge}} measures the number of overlapping n-grams between the generated hypotheses and a set of gold references.
\end{itemize}
For the n-gram-based metrics, we use the sacrebleu~\citep{sacrebleu}\footnote{https://github.com/mjpost/sacrebleu} toolkit to perform the calculations. 

\paragraph{Neural-based metrics.}
For WMT22 Metric Shared Task, we list the results of most metrics in the task, including BERTscore~\citep{Zhang*2020BERTScore:}, MATEESE-QE~\citep{perrella-etal-2022-matese}, MS-COMET-QE-22~\citep{kocmi2022ms}, UNITE-src~\citep{wan2022alibaba}, COMET-QE~\citep{rei-etal-2022-comet}, COMETKiwi~\citep{rei-etal-2022-comet}, YiSi-1~\citep{lo-2019-yisi}, MATESE~\citep{perrella-etal-2022-matese}, MS-COMET-22~\citep{kocmi2022ms}, UniTE~\citep{wan2022alibaba}. We mainly conduct experiments on two most widely used and powerful metrics:
\begin{itemize}
    \item \textbf{BLEURT~\citep{sellam2020bleurt}} is a learned metric fine-tuned to directly assess given translations by jointly encoding them with their references. We use the BLEURT20 checkpoint.
    \item \textbf{COMET-20~\citep{rei-etal-2020-comet}} is a learned metric fine-tuned to provide a z-standardized score for given translations. It compares their representations to source and reference embeddings. We use the default wmt20-comet-da model from v1.1.2. We did not use COMET-22 as its ensemble nature and lack of open-source models.
\end{itemize}
For BLEURT\footnote{https://github.com/google-research/bleurt} and COMET\footnote{https://github.com/Unbabel/COMET}, we use the official github example scripts.

We also report the result of GEMBA~\citep{kocmi2023large} which use Davinci-003 to direct assess the model outputs.

\subsection{Settings}
We use gpt-3.5-turbo for reference generation. When calling the OpenAI API, we use the default values for all hyper-parameters. We generate 40 references for each sentence on WMT22 Metric Shared Tasks as we conduct more analyses on this and 10 references on the SummEval benchmark. 

\begin{table}[t]
\begin{center}
\scalebox{0.63}{
\begin{tabular}{ll|c|c|c|c}
\toprule
\multirow{2}{*}{\textbf{ID}} & \multirow{2}{*}{\textbf{Metrics}} & \multirow{2}{*}{\textbf{Accuracy}}  & \multicolumn{1}{c|}{\textbf{en-de}} & \multicolumn{1}{c|}{\textbf{en-ru}} & \multicolumn{1}{c}{\textbf{zh-en}} \\ 
& & & \textbf{$\rho$} & \textbf{$\rho$} & \textbf{$\rho$} \\
\midrule
0 & GEMBA & 88.0\% & - & -  & -  \\
1 & \bf BLEURT-20-LLM-Ref & 86.4\% & 0.726 & 0.962 & 0.938 \\
2 & \bf COMET-20-LLM-Ref & 85.4\% & 0.864 & 0.946 & 0.963 \\
3 & MetricX XXL & 85.0\% & 0.847 & 0.949 & 0.938 \\
4 & BLEURT-20 & 84.7\% & 0.719 & 0.959 & 0.938 \\
5 & COMET-22 & 83.9\% & 0.771 & 0.900 & 0.942 \\
6 & COMET-20 & 83.6\% & 0.876 & 0.936 & 0.970 \\
7 & UniTE & 82.8\% & 0.624 & 0.888 & 0.914 \\
8 & MS-COMET-22 & 82.8\% & 0.695 & 0.809 & 0.909 \\
9 & \bf f200spBLEU-LLM-Ref-DAS & 81.3\% & 0.427 & 0.875 & 0.923 \\
10 & MATESE & 81.0\% & 0.617 & 0.757 & 0.856 \\
11 & \bf f200spBLEU-LLM-Ref & 79.6\% & 0.424 & 0.851 & 0.869 \\
12 & YiSi-1 & 79.2\% & 0.626 & 0.881 & 0.935\\
13 & COMETKiwi[noref] & 78.8\% & 0.674 & 0.763 & 0.866 \\
14 & \bf chrF-LLM-Ref  & 78.4\% & 0.587 & 0.878 & 0.861 \\
15 & COMET-QE[noref] & 78.1\% & 0.502 & 0.468 & 0.569 \\
16 & BERTScore & 77.4\% & 0.428 & 0.811 & 0.924 \\
17 & \bf BLEU-LLM-Ref  & 76.6\% & 0.378 & 0.784 & 0.783 \\
18 & UniTE-src[noref] & 75.9\% & 0.509 & 0.779 & 0.874 \\
19 & MS-COMET-QE-22[noref] & 75.5\% & 0.539 & 0.672 & 0.897 \\
20 & MATESE-QE[noref] & 74.8\% & 0.337 & 0.637 & 0.767 \\
21 & f200spBLEU & 74.1\% & 0.283 & 0.819 & 0.728 \\
22 & chrF & 73.3\% & 0.346 & 0.815 & 0.630 \\
23 & BLEU & 70.8\% & 0.179 & 0.724 & 0.594 \\
\bottomrule
\end{tabular}
}
\caption{System-level Pearson correlation (\textbf{$\rho$}) and \textbf{Accuracy} for WMT22 Metrics Shared Task. The results in the table are ranked by \textbf{Accuracy}. The \textbf{Bolded} metrics are our results. The metrics with \textbf{LLM-Ref} suffix represent that the metrics use multiple references constructed by our method. The metrics with \textbf{DAS} suffix denotes Diversity-Aware Selection.}
\label{tab:system}
\end{center}
\end{table}

\begin{table}[t]
\begin{center}
\scalebox{0.63}{
\begin{tabular}{ll|c|c|c|c}
\toprule
\multirow{2}{*}{\textbf{ID}} & \multirow{2}{*}{\textbf{Metrics}} & \multirow{2}{*}{\textbf{Accuracy}}  & \multicolumn{1}{c|}{\textbf{en-de}} & \multicolumn{1}{c|}{\textbf{en-ru}} & \multicolumn{1}{c}{\textbf{zh-en}} \\ 
& & & \textbf{$\tau$} &\textbf{$\tau$} &\textbf{$\tau$} \\
\midrule
0 & GEMBA & 88.0\% & 0.310 & 0.330 & 0.370 \\
1 & \bf BLEURT-20-LLM-Ref & 86.4\% & 0.342 & 0.391 & 0.386 \\
2 & \bf COMET-20-LLM-Ref & 85.4\% & 0.323 & 0.374 & 0.367 \\
3 & MetricX XXL & 85.0\% & 0.360 & 0.420 & 0.427 \\
4 & BLEURT-20 & 84.7\% & 0.344 & 0.359 & 0.361 \\
5 & COMET-22 & 83.9\% & 0.368 & 0.400 & 0.428 \\
6 & COMET-20 & 83.6\% & 0.319 & 0.330 & 0.332 \\
7 & UniTE & 82.8\% & 0.369 & 0.378 & 0.357 \\
8 & MS-COMET-22 & 82.8\% & 0.283 & 0.351 & 0.341 \\
9 & \bf f200spBLEU-LLM-Ref-DAS & 81.3\% & 0.220 & 0.233 & 0.224 \\
10 & MATESE & 81.0\% & 0.323 & 0.279 & 0.389 \\
11 & \bf f200spBLEU-LLM-Ref & 79.6\% & 0.231 & 0.246 & 0.228 \\
12 & YiSi-1 & 79.2\% & 0.235 & 0.227 & 0.296 \\
13 & COMETKiwi[noref] & 78.8\% & 0.290 & 0.359 & 0.364 \\
14 & \bf chrF-LLM-Ref & 78.4\% & 0.246 & 0.256 & 0.216 \\
15 & COMET-QE[noref] & 78.1\% & 0.281 & 0.341 & 0.365 \\
16 & BERTScore & 77.4\% & 0.232 & 0.192 & 0.316 \\
17 & \bf BLEU-LLM-Ref & 76.6\% & 0.205 & 0.214 & 0.222 \\
18 & UniTE-src[noref] & 75.9\% & 0.287 & 0.342 & 0.343 \\
19 & MS-COMET-QE-22[noref] & 75.5\% & 0.233 & 0.305 & 0.287 \\
20 & MATESE-QE[noref] & 74.8\%& 0.244 & 0.229 & 0.337 \\
21 & f200spBLEU & 74.1\% & 0.180 & 0.153 & 0.140 \\
22 & chrF & 73.3\% & 0.214 & 0.168 & 0.147 \\
23 & BLEU & 70.8\% & 0.169 & 0.140 & 0.145 \\
\bottomrule
\end{tabular}
}
\caption{Segment-level Kendall correlation (\textbf{$\tau$}) for WMT22 Metrics Shared Task. The results in the table are ranked by \textbf{Accuracy}. The \textbf{Bolded} metrics are our results. The metrics with \textbf{LLM-Ref} suffix represent that the metrics use multiple references constructed by our method. The metrics with \textbf{DAS} suffix denotes Diversity-Aware Selection.}
\label{tab:segment}
\end{center}
\end{table}

\begin{table}[t]
\begin{center}
\scalebox{0.61}{
\begin{tabular}{l|c|c|c|c}
\toprule
\textbf{Metrics} & \textbf{Coherence} & \textbf{Consistency} & \textbf{Fluency} & \textbf{Relevance} \\ 
\midrule
ROUGE-1 & 0.1215 & 0.1588 & 0.1067 & 0.2561 \\
ROUGE-2 & 0.0986 & 0.1706 & 0.1071 & 0.1850 \\
ROUGE-L & 0.1040 & 0.1439 & 0.1074 & 0.2400 \\
\midrule
\textbf{ROUGE-1-LLM-Ref} & \textbf{0.1843} & 0.1597 & 0.1063 & 0.2533 \\
\textbf{ROUGE-2-LLM-Ref} & 0.1648 & \textbf{0.1835} & \textbf{0.1235} & \textbf{0.2843} \\
\textbf{ROUGE-L-LLM-Ref} & 0.1201 & 0.1612 & 0.1251 & 0.2419 \\
\bottomrule
\end{tabular}
}
\caption{Spearman score of sample-level correlation one SummEval benchmark. The \textbf{Bolded} metrics are our results. The metrics with \textbf{LLM-Ref} suffix represent that the metrics use multiple references constructed by our method.}
\label{tab:summ}
\end{center}
\end{table}

\subsection{Results on WMT22 Metrics Shared Task}
We evaluated the performance of LLM-Ref at both the system level and segment level. 

\paragraph{System-level Performance.}
Table \ref{tab:system} presents the pairwise accuracy and Pearson correlation at the system level. LLM-Ref demonstrates positive improvements across all metrics, including both neural-based and n-gram-based metrics. The improvement is particularly noticeable in the n-gram-based metrics, with an accuracy increase of up to \textbf{+5.5} percentage points when comparing ID 21 to 11. When combined with Diversity-Aware selection for reference candidates, there is an improvement of up to \textbf{+7.2} percentage points from ID 21 to 9.

For the neural-based metrics, both BLEURT and COMET-20 exhibit improvements of approximately \textbf{+1.7} percentage points comparing ID 6 and 4 to 2 and 1. Moreover, after incorporating LLM-Ref, BLEURT and COMET-20 outperform all metrics except GPT-3.5, establishing them as the non-LLM state-of-the-art (SOTA) metrics.

Furthermore, in terms of the Pearson correlation (\textbf{$\rho$}), the improvement is even more pronounced in the n-gram-based metric, with F200spBLEU experiencing an improvement of up to \textbf{+19.5} percentage points one ZH-EN from ID 21 to 9. The improvement in the neural-based metric is relatively tiny, with a maximum increase of 1 percentage point.

\paragraph{Segment-level Performance.}
We also examined the segment-level performance of LLM-Ref, considering that GEMBA's performance at the segment level is not the best and lacks stability. The Kendall-Tau (\textbf{$\tau$}) results for each language pair are presented separately in Table \ref{tab:segment}. LLM-Ref exhibits significant improvements in Kendall-Tau (\textbf{$\tau$}) for almost all metrics. With the integration of LLM-Ref, BLEURT and COMET-20 achieve maximum increases of \textbf{+3.2} and \textbf{+3.5} percentage points comparing ID 6, 4 to 2, 1, respectively. Furthermore, F200spBLEU demonstrates an improvement of up to \textbf{+9.3} percentage points from ID 21 to 11 on EN-RU.

It is worth noting that GEMBA does not excel in segment-level performance even lower than BLEURT and COMET-20, which suggests that LLMs are not good enough at the segment-level evaluation. However, the reference candidates generated by LLMs still have consistent improvements on solid baselines.

\subsection{Results on SummEval benchmark}
We mainly conducted experiments using the ROUGE metric on the SummEval. As shown in Table \ref{tab:summ}, LLM-Ref also consistently improves performance across the four domains of the summarization task. Specifically, on the Relevance, the ROUGE-2 improves by +9.93 points, indicating substantial gains.

\begin{figure*}[t!]
\begin{center}
     \scalebox{0.9}{
      \includegraphics[width=1\textwidth]{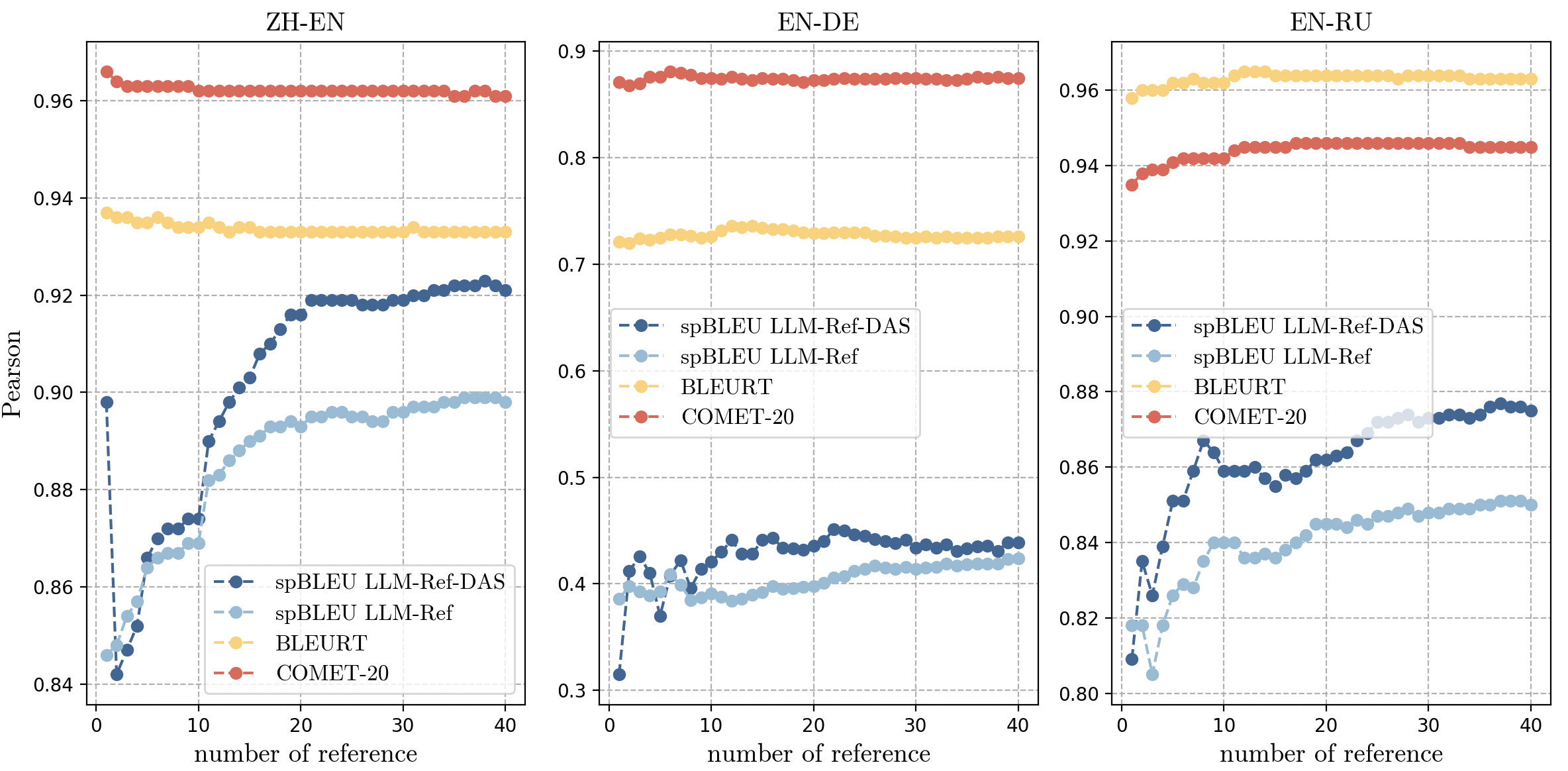}
      } 
      \caption{
      The system-level Pearson correlation (\textbf{$\rho$}) for each metric along with the increasing reference numbers. The growth of N-gram-based metrics is more pronounced.
      } 
      \label{fig:ref_num}  
 \end{center} 
\end{figure*}

\section{Analysis}
In this section, we will first analyze the effects of reference numbers and prompt variants, and then analyze whether multi-reference solves the data leakage issue. We conduct these analyses on WMT22 Metrics Shared Task.

\subsection{Effects of Reference Numbers}
Although LLMs are capable of generating a large number of references, it is important to consider the impact of reference numbers on their effectiveness, as the diversity of the generated reference may decrease.

\paragraph{System-level Pearson.}
Figure \ref{fig:ref_num} illustrates the variation in system-level Pearson correlation with increasing reference numbers for the three translation directions. For n-gram-based metrics, the Pearson correlation generally tends to increase as the number of references increases. This increase is more evident for Chinese-English and English-Russian, while the increase is slower for English-German. 
Furthermore, incorporating Diversity-Aware Selection boosts the Pearson correlation considerably for the three directions. Notably, the impact of the ground truth on the system-level performance of n-gram-based metrics is consistently negative across all languages so we only use the generated references.

For the neural network metrics, the Pearson correlation exhibits a decrease in ZH-EN as the number of references increases, while it increases for EN-DE and EN-RU. However, for EN-DE, the Pearson correlation starts to decrease after a certain number of references (around 6-10). We assume that since BLEURT and COMET-20 already have strong performance (0.97 on ZH-EN and 0.95 on EN-RU), some low-quality reference candidates may be more unbeneficial. In addition, neural network metrics only calculate scores based on a single reference and then perform a basic max operation, which constrains their capability to use multiple references.

\begin{figure*}[t!]
\begin{center}
     \scalebox{0.9}{
      \includegraphics[width=1\textwidth]{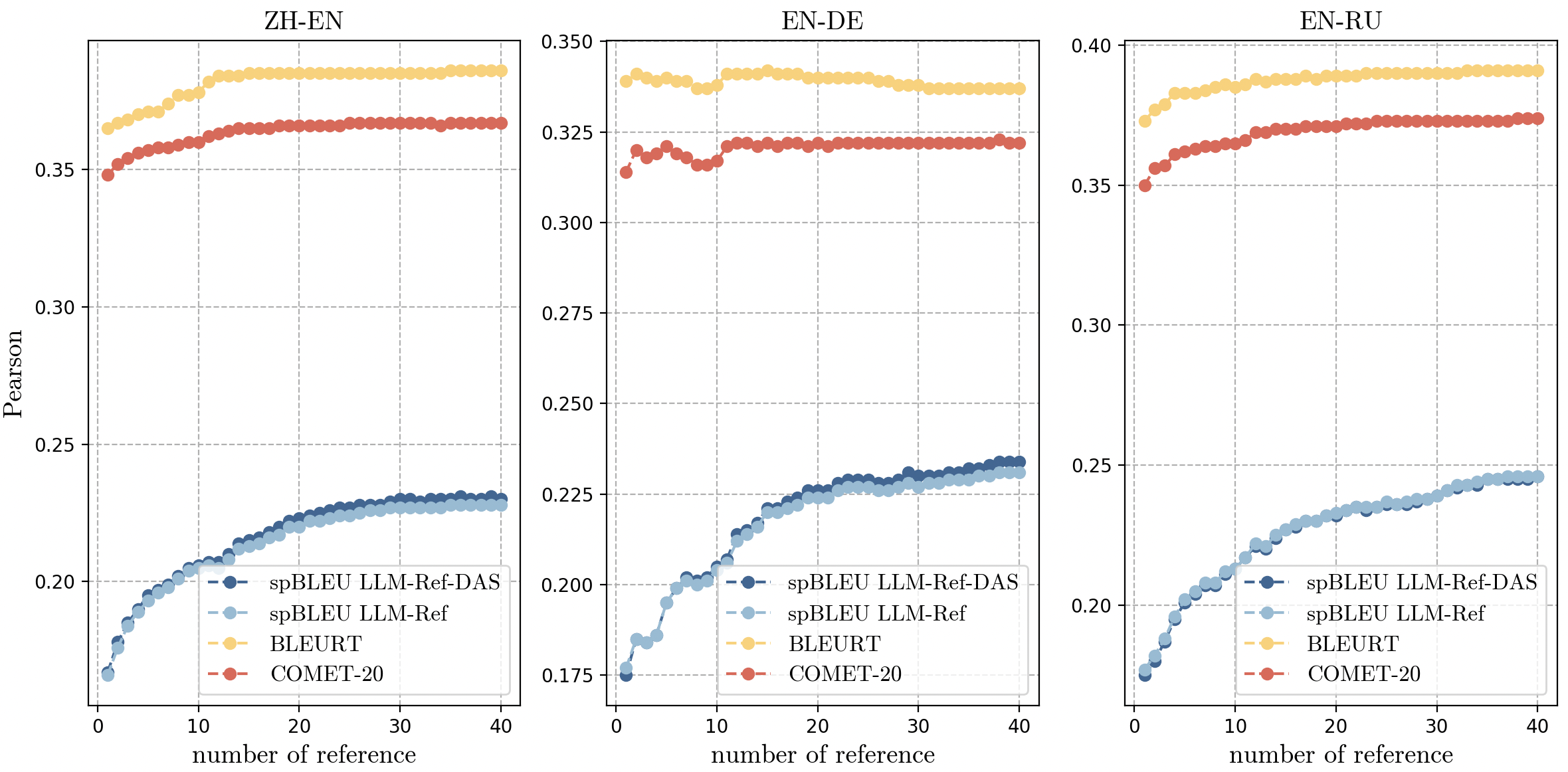}
      } 
      \caption{
      The segment-level Kendall correlation (\textbf{$\tau$}) for each metric along with the increasing reference numbers. Suffix with \textbf{GT} indicates the use of ground truth. Most of the metrics have a stable gain with increasing reference numbers.
      } 
      \label{fig:ref_num_seg}  
 \end{center} 
\end{figure*}

\paragraph{Segment-level Kendall-tau.}
Figure \ref{fig:ref_num_seg} presents the Kendall-Tau correlation (\textbf{$tau$}) at the segment level. 
The correlation increases more consistently with an increasing number of references in the three directions, which shows an upward trend at around 40 references.
Diversity-Aware Selection shows slight improvements in the segment level.
In contrast to the system level, we observe that the effect of ground truth on n-gram-based metrics is consistently positive so we use it during the segment-level evaluation. 

Additionally, the overall relevance of neural network metrics to human evaluation also improves along with the increasing number of references. The improvement is relatively smaller for neural network metrics compared to n-gram-based metrics due to their higher baseline performance.

\begin{table}[t]
\begin{center}
\scalebox{0.82}{
\begin{tabular}{ll|c|c}
\toprule
\textbf{ID} & \multirow{1}{*}{\textbf{Prompt Variant}} & \multicolumn{1}{c|}{\textbf{(system) $\rho$}} & \textbf{(segment) $\tau$} \\ 
\midrule
0 & Ground Truth & 0.728 & 0.140 \\
\midrule
1 & MT system & 0.537 & 0.180 \\
\midrule
\multicolumn{4}{c}{Reference Candidates Generated By LLM} \\
\midrule
2 & Chinese Prompt& 0.780 & 0.196 \\
3 & English Prompt& 0.793 & 0.193 \\
4 & 3 + No Ground Truth & 0.693 & 0.193 \\
5 & 3 + No Rules & 0.791 & 0.193 \\
\bottomrule
\end{tabular}
}
\caption{The impact of different prompts on the quality of the generated translations. The experiments were conducted on the ZH-EN test set.}
\label{tab:prompt}
\end{center}
\end{table}

\subsection{Effects of Prompt Variants}
\label{sec:prompt}
We have investigated the impact of different prompts on the quality of the generated reference candidates. We also try to use commercial translation systems (e.g. Google, Microsoft) to generate references. The results are summarized in Table \ref{tab:prompt}. All experiments on prompt variants were conducted on the ZH-EN dataset and the number of generated reference candidates is constrained to 2.

Reference candidates generated by commercial translation systems do significant harm to system-level Pearson correlation (\textbf{$\rho$}), implying that commercial translation systems may not be able to produce high-quality references to support multiple references. 
Regarding the choice of language for the prompt, the reference candidates generated by the English prompt slightly outperform the ones generated by the Chinese prompt. As ground truth is optional in the prompt, we find omitting the ground truth from the prompt leads to a considerable drop in the quality of the generated references, resulting in a 10 percentage point decrease in the Pearson correlation (\textbf{$\rho$}) of the evaluation metric. The rules (characterization and task-specific rules) in the prompt do not appear to have much impact. 
The Kendall correlation (\textbf{$\tau$}) at the segment level does not vary noticeably across methods.

\subsection{Generate As Many References As Possible At Once}
\label{sec:ref_num}
\begin{figure}[t!]
\begin{center}
     \scalebox{0.9}{
      \includegraphics[width=0.45\textwidth]{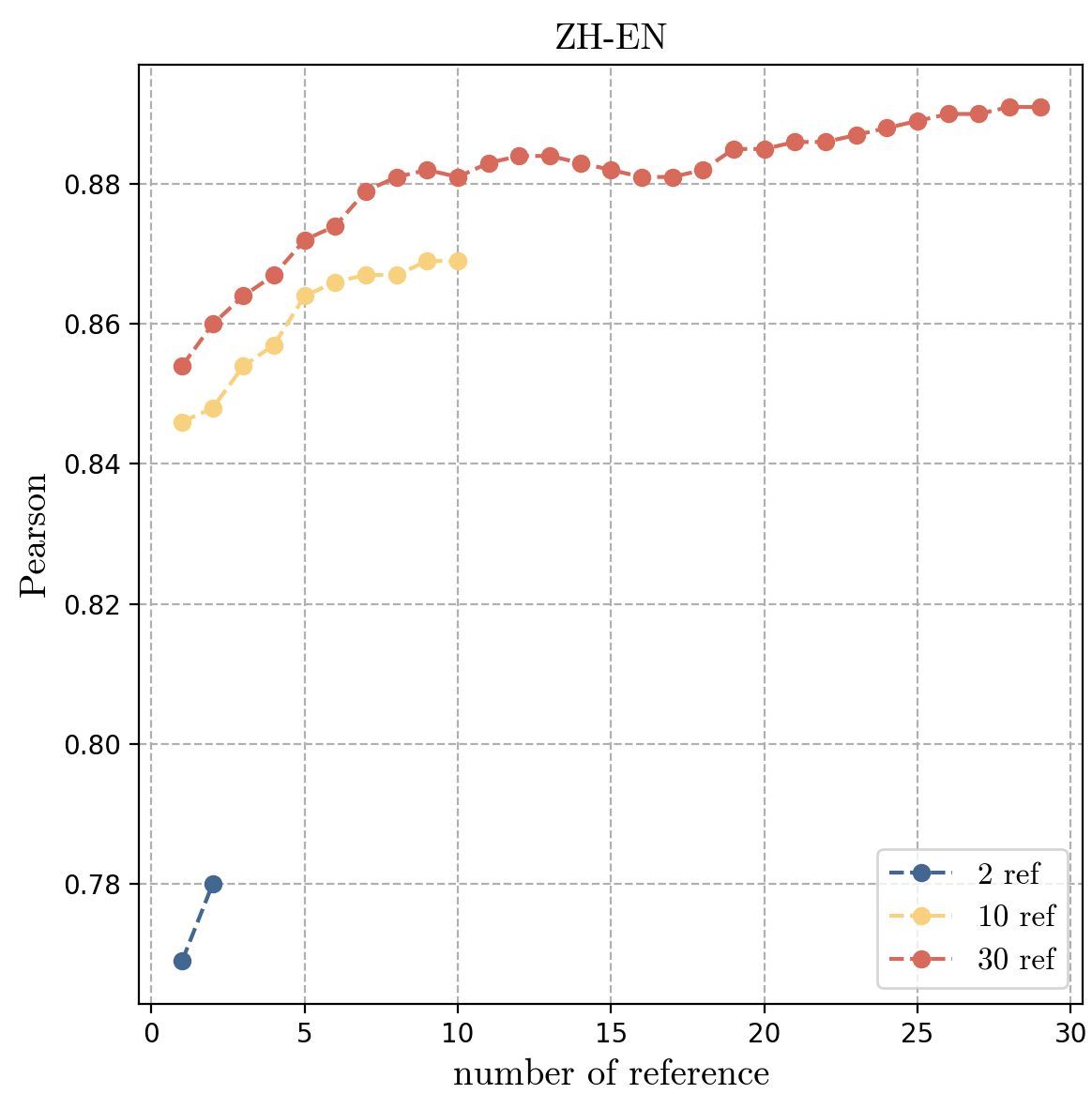}
      } 
      \caption{
      Pearson correlation of F200spBLEU with different reference numbers in a single inference. Generating more references in a single inference cause higher-quality references. 
      } 
      \label{fig:more_ref_gene}  
 \end{center} 
\end{figure}

It is worth noting that the number of generated translations specified in the prompt has a significant effect on the quality of the results. As shown in Figure \ref{fig:more_ref_gene}. increasing the number of references from 2 to 30 leads to an 8 percentage point increase in the Pearson correlation of F200spBLEU. 
We speculate that as LLM generates more reference candidates, it strives to produce reference candidates at a finer granularity, which enables more distinct discrimination between high and low-quality candidates. Accordingly, the general quality of the reference candidatess improves.

\begin{table}[t]
\begin{center}
\scalebox{0.75}{
\begin{tabular}{l|cc|cc}
\toprule
\multirow{2}{*}{\textbf{Model}} & \multicolumn{2}{c|}{\textbf{Gold-Ref}} & \multicolumn{2}{c}{\textbf{Multi-Ref}}  \\
& spBLEU & BLEURT & spBLEU & BLEURT \\
\midrule
\multicolumn{5}{c}{Large Language Models} \\
\midrule
BLOOMz-7b-ft & \textbf{34.46} & \textbf{69.33} & 49.05 & 73.85 \\
Claude & 22.16 & 68.23 & 49.50 & 74.02 \\
GPT3.5 & 22.94 & 68.45 & \textbf{52.37} & \textbf{74.16} \\
\midrule
\multicolumn{5}{c}{Machine Translation models} \\
\midrule
MT & 27.05 & 69.18 & 52.76 & 74.41 \\
MT-ft-test & 35.86 & 71.70 & 53.08 & 75.73 \\
\bottomrule
\end{tabular}
}
\caption{F200spBLEU and BLEURT score on Flores200 test set. The \textbf{Bolded} scores correspond to the best in LLMs. MT-ft-test is the MT model finetuned on the test set to simulate the data leakage issue.}
\label{tab:flores_multi}
\end{center}
\end{table}

\subsection{Do Multi-Reference Solve Data Leakage Issue}
As previously discussed, data leakage can be a concern when using open-source LLMs, as it is difficult to determine the exact data processing methods used during pre-training, and we cannot guarantee that the models have not seen the data from the test set. Therefore, we aim to explore whether multiple references can alleviate this problem. We verify this on the Flores200 Japanese-Chinese test set~\citep{costa2022no} as in preliminary experiments.

\paragraph{N-gram Based Metrics.}
The results are presented in Table \ref{tab:flores_multi}. 
When using a single reference, BLOOMz-7b-ft exceeds the strong MT model of +6.95 and even closed-source LLMs with several hundred billion parameters (Claude and GPT3.5) up to +12.3 due to data leakage issues.
After using multiple references, the MT model and other closed-source LLMs in turn outperform BLOOMz-7b-ft. GPT3.5 also achieves comparable results to the MT model.

To demonstrate that the model's performance is accurately measured by multiple references, we finetune the MT model on the test set to simulate the data leakage problem. It can be observed that the MT model after finetuning performs significantly better with a single reference, exceeding +8.81 F200spBLEU compared to the performance before finetuning. However, when multiple references are used, the performance gap drops to +0.32 F200spBLEU, aligning with our expectations as the base model is the same one. This result indicates that multiple references can effectively mitigate the impact of data leakage.

\paragraph{Neural Network Based Metrics.}
In contrast to n-gram-based metrics, we find that neural network based metrics have limited ability to mitigate data leakage when multiple references are used. The difference in BLEURT between MT-ft-test and MT decreases from +2.52 to +1.32 after incorporating multiple references, which is still a noticeably large gap in BLEURT scores.

We speculate that neural network based metrics rely on the semantic space to determine the similarity between translations. As long as the generated translation is semantically similar to the added references, it will receive a higher score. This inability to discriminate and penalize overfitting results in terms of diversity constrains the capability of neural network metrics in mitigating data leakage issue.

Based on these observations, we believe that the use of n-gram-based metrics is essential for the subsequent evaluation of NLG tasks, especially when evaluating LLMs. To ensure the validity of the evaluations, employing multiple references with n-gram-based metrics can effectively mitigate potential data leakage issues and minimize misevaluation.

\section{Related Work}
\subsection{Automatic Evaluation Metrics}
Automated evaluation metrics can be divided into two categories based on the algorithm used, n-gram-based metrics and neural network metrics. The former computes the corresponding accuracy by matching the results generated by the model with the manually annotated reference at the character level of the n-gram, e.g., BLEU~\citep{papineni2002bleu}, CHRF~\citep{popovic-2015-chrf}, ROUGE~\citep{lin-2004-rouge}, and METEOR~\citep{banerjee-lavie-2005-meteor}. 
The latter measures the similarity of sentences by learning on a large amount of text data with high-dimensional semantic vectors, e.g., BLEURT~\citep{sellam2020bleurt}, COMET~\citep{rei-etal-2022-comet}, and BERTScore~\cite{Zhang*2020BERTScore:}. As neural network based metrics have high agreements with human assessment, it is increasingly being used for automated assessment. 

\subsection{LLM as an Evaluator}
The use of LLMs as evaluators has gained popularity in recent studies~\citep{chiang2023can, wang2023chatgpt}. This approach leverages the strong generalizability of models such as GPT4, GPT3.5, and ChatGPT to score the results of various NLG tasks. It has shown high agreement with human evaluation and has been applied to tasks such as machine translation, summarization, and image captioning.

Previous work in this area typically involves directly invoking the LLMs through prompts to score the task outcomes without generating corresponding references to enhance the efficacy of other metrics. The approach closest to ours is one that exploits the paraphrase ability of LLMs to generate multiple references to augment the consistency of automatic metrics ~\citep{tang2023not}. A key difference is that they do not distinguish between the generated candidates whereas we propose a Diversity-Aware strategy to select references with high diversity. In addition, we emphasize that multiple references together with n-gram-based metrics can alleviate the data leakage issue, which they do not.

\section{Conclusion}
In this paper, we propose LLM-Ref, a framework that enhances the evaluation of NLG tasks by leveraging LLMs. By generating multiple reference candidates and implementing a Diversity-Aware selection mechanism, we effectively address the limited diversity and data leakage challenges associated with NLG evaluations. Experimental results demonstrate that LLM-Ref significantly improves the consistency of automatic evaluation metrics with human assessments. Additionally, our analyses highlight the efficacy of n-gram-based metrics with multiple references in mitigating data leakage risk in LLMs while neural network metrics struggle to overcome.  

\bibliography{anthology,custom}
\bibliographystyle{acl_natbib}

\end{document}